\documentclass[runningheads]{llncs}
\RequirePackage{fix-cm}
\usepackage{comment}
\usepackage{graphicx}
\usepackage{amsfonts,amssymb}
\usepackage{amsmath}
\usepackage{multirow}
\usepackage{booktabs}
\usepackage{latexsym}
\usepackage[misc]{ifsym}

\begin{document}
\title{Nested Named Entity Recognition from Medical Texts: An Adaptive Shared Network Architecture with Attentive CRF}

\def\CICAISubNumber{139}  

\titlerunning{Nested Named Entity Recognition from Medical Texts}

\author{Junzhe Jiang\inst{1,2}\and
Mingyue Cheng\inst{1,2}\and
Qi Liu\inst{1,2,*}\and
Zhi Li\inst{3}\and
Enhong Chen\inst{1,2}
}
\authorrunning{J. Jiang et al.}
%
\institute{Anhui Province Key Laboratory of Big Data Analysis and Application, University of Science and Technology of China, Hefei, China\and State Key Laboratory of Cognitive Intelligence, Hefei, China\and Shenzhen International Graduate School, Tsinghua University, Shenzhen, China
\email{\{jzjiang,mycheng\}@mail.ustc.edu.cn, \{qiliuql,cheneh\}@ustc.edu.cn, zhilizl@sz.tsinghua.edu.cn}}
\maketitle              

\begin{abstract}
Recognizing useful named entities plays a vital role in medical information processing, which helps drive the development of medical area research. Deep learning methods have achieved good results in medical named entity recognition (NER). However, we find that existing methods face great challenges when dealing with the nested named entities. In this work, we propose a novel method, referred to as ASAC, to solve the dilemma caused by the nested phenomenon, in which the core idea is to model the dependency between different categories of entity recognition. The proposed method contains two key modules: the adaptive shared (AS) part and the attentive conditional random field (ACRF) module. The former part automatically assigns adaptive weights across each task to achieve optimal recognition accuracy in the multi-layer network. The latter module employs the attention operation to model the dependency between different entities. In this way, our model could learn better entity representations by capturing the implicit distinctions and relationships between different categories of entities. Extensive experiments on public datasets verify the effectiveness of our method. Besides, we also perform ablation analyses to deeply understand our methods.

\end{abstract}
\section{Introduction}

\begin{figure}[tbhp]
\center
\includegraphics[width=0.6\textwidth]{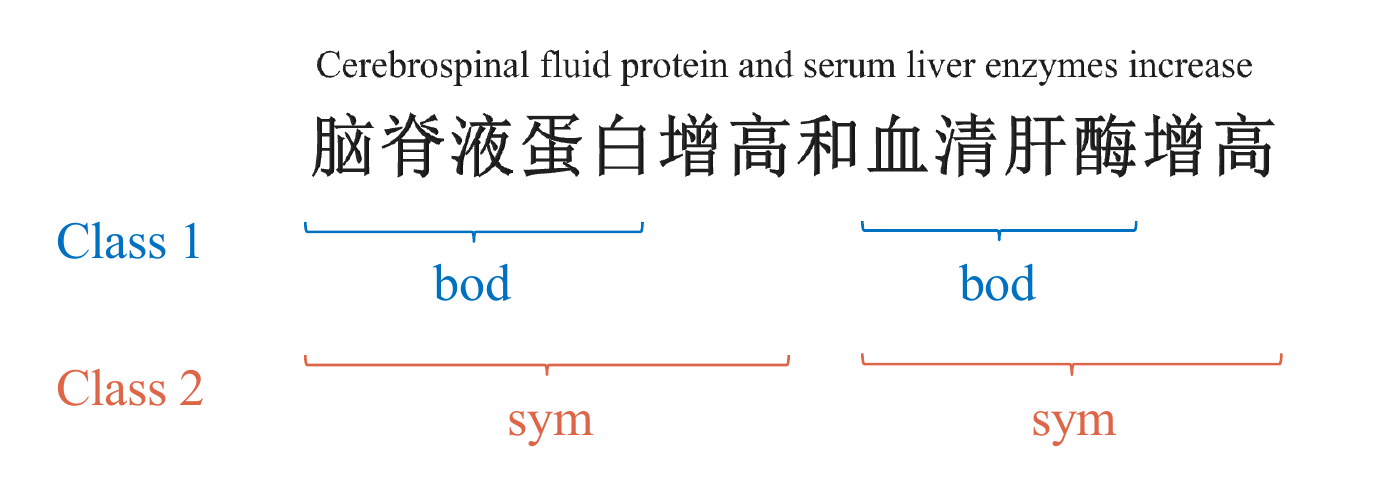}
\caption{An example of nested NER in Chinese medical texts, while ‘bod’ and ‘sym’ represent \textit{body} and \textit{symptom} entities, respectively.} \label{fig1}
\end{figure}

Natural language documents in the medical field, such as medical textbooks, medical encyclopedias, clinical cases, and test reports, contain much medical expertise and terminology. The key idea of understanding medical data is to extract critical knowledge from the medical text accurately~\cite{liu2022chinese}. Therefore, accurate and rapid extraction of medical entities and transformation of these unstructured data into structured domain knowledge graphs are crucial for obtaining and exploiting medical information. Named entity recognition (NER) based on deep learning applies machines to read medical texts, which significantly improve the efficiency and quality of medical research and serve downstream subtasks~\cite{li2020survey}.

However, there are often cases of nested named entities in texts in the medical field. Figure~\ref{fig1} displays an example of nested NER in Chinese medical texts, where the \textit{symptom} named entity contains the \textit{body} named entity. The foremost sequence labeling method~\cite{huang2015bidirectional,akbik2018contextual} is only valid for non-nested entities (or flat entities). Previous studies have given some solutions, which regard NER either as question answering~\cite{mengge2020coarse}, span classification~\cite{jiang-etal-2020-generalizing}, dependency parsing tasks~\cite{yu-etal-2020-named}, or discrete joint model~\cite{ji2021neural}. However, despite the success of span prediction based systems, these methods suffer from some different weaknesses~\cite{shen2021locate}. First, their method suffers from the boundary inconsistency problem due to the separate decoding procedures. Then, due to numerous low-quality candidate spans, these methods require high computational costs. Next, it is hard to identify long entities frequently found in medical texts because the length of the span enumerated during training is not infinite.

Different from the above researches, layered methods solve this task through multi-level sequential labeling~\cite{shibuya2020nested,fisher2019merge}, which can capture the dependencies between adjacent word-level labels and maximize the probability of predicted labels over the whole sentence~\cite{wang2022nested}. However, these methods always simply divide entities into several levels, where the term level indicates the depth of entity nesting, and sequential labeling is performed repeatedly. Therefore, the differences between entity categories have not been noticed, since just identifying the results of each layer independently. Moreover, there are often implicit relationships between the recognition results of different entity categories, such as \textit{symptom} and \textit{body}, but it is easy to be ignored or only pass results from lower layers to upper layers without the reverse way.

In this paper, we propose a \textbf{A}daptive \textbf{S}hared network architecture with \textbf{A}ttentive \textbf{C}onditional random fields namely \textbf{ASAC} based on the end-to-end sequence labeling, which can handle the nested NER for medical texts. Our model not only takes advantage of sequence labeling, but also equally considers the implicit distinctions and relationships between entities in different layers. More specifically, we introduce the adaptive shared mechanism for pre-trained language model at encoding, to adaptively learn the difference between different entities recognition tasks~\cite{liu2019exploiting}. This method has a better fitting ability than the hard-shared method~\cite{sun2020adashare,cheng2021learning}, which shares different modules according to a fixed strategy. Besides, we utilize attentive conditional random fields to explore the relationships among the multi-layer learning results at decoding based on attention mechanism~\cite{liang2020abstractive}.

Our main contributions are as follows:

\begin{itemize}
    \item We introduce a novel ASAC model based on deep learning methods, which can efficiently extract named entities from medical texts. We leverage effective mechanisms to enable the model to adaptively discover distinctions and connections between multiple entity recognition tasks in parallel, in order to enhance the performance of the NER task.

    \item We adopt the Adaptive Shared (AS) mechanism~\cite{sun2020adashare} to adaptively select the output of each layer of the pre-trained model for encoding the input texts, thus obtaining the different characteristics of different entity categories. Through this mechanism, we can learn the contextual features from information contained in different layers of the pre-trained language model, for the downstream corresponding tasks.

    \item We exploit the Attentive Conditional Random Fields (ACRF) model in the decoding stage to get the labels. It can use the \textit{Viterbi} decoding outputs of the other entity recognition tasks as a query. Then, input the query through the attention mechanism as the residual to the origin CRF for deviation correction. In this way, the recognition results of other levels can be integrated to improve the recognition effect of nested named entities.
    
    \item We evaluate our proposed model on a public dataset of Chinese medical NER, namely the Chinese Medical Entity Extraction dataset~\cite{hongying2020building} (CMeEE). Extensive experiment results show the effectiveness of the proposed model.
\end{itemize}

\section{Related Work}
In this section, we review the related work on the current approaches for NER, including the pre-trained language model, nested NER algorithm, and the Chinese NER in the medical domain.

\subsection{Chinese Medical NER}
With the increasing demand for NER in the medical field, there have emerged some related medical competitions and datasets, such as CHIP\footnote{http://www.cips-chip.org.cn/} and CCKS\footnote{http://sigkg.cn/ccks2022/}. However, due to privacy or property rights protection issues, the medical NER corpora, especially the Chinese medical NER corpora, is particularly scarce~\cite{liu2022chinese}. There is no other officially authorized way for these datasets. So most of the research on Chinese medical NER is based on China Conference on Knowledge Graph and Semantic Computing (CCKS) datasets.

When it comes to solutions, Chinese NER methods based on deep learning have gradually become dominant and have achieved continuous performance improvements~\cite{kou2018semantic}. Cai et al.~\cite{cai2019deep} added attention layers between the character representation and the context encoder. Zhu et al.~\cite{zhu2019can} used a convolutional attention layer between the feature representation and the encoder.

\subsection{Nested NER}
Previously, some researchers have also discussed the nested NER based on deep learning. There are various solutions, mainly including the following:
(a) transform the decoding process into multi-classification decoding~\cite{strakova2019neural,li2020learning}.
(b) span-based methods which treat NER as a classification task on a span with the innate ability to recognize nested named entities~\cite{fu2021spanner,jiang-etal-2020-generalizing}.
(c) use other modeling methods instead of sequence labeling and span-based methods, such as machine reading comprehension task~\cite{li2020unified}, constructing a hyper-graph~\cite{wang2018neural}, etc

\subsection{Pre-trained Model}
In recent years, the emergence of the pre-trained model has brought NLP into a new era. Many researches~\cite{liu2022chinese} have shown that the pre-trained model trained on a very large corpus can learn a lot of language text representation suitable for different domains. This approach can often help different downstream NLP tasks, including NER. Among various well-known pre-trained language models, the BERT model~\cite{devlin2018bert} with excellent results is widely used. BERT is a bi-directional encoder based on transformer. It can mine the representation information contained in the context through the self-supervised learning task. Some researchers have carried out more research on the basis of BERT and put forward many improved models, such as RoBERTa~\cite{liu2019roberta}, AlBERT~\cite{lan2019albert}, and so on.

\section{Methodology}
\begin{figure}
\includegraphics[width=\textwidth]{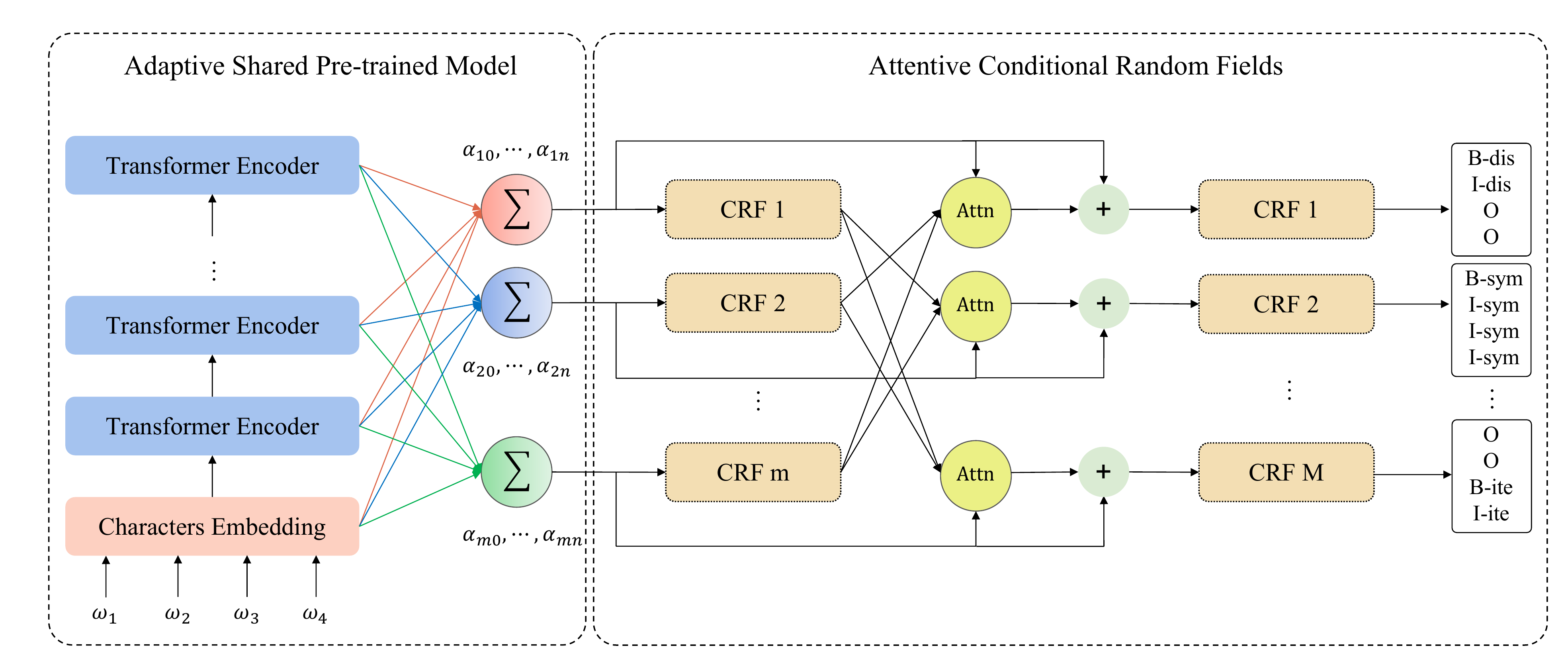}
\caption{The architecture of our model. The dotted lines mean these components are shared across levels. $\alpha_{ij}$ denotes the adaptive weight, $Attn$ represents the attention mechanism.} \label{fig2}
\end{figure}

Our ASAC model can be divided into two modules. The former part is an adaptive shared pre-trained language model for encoding the input texts to capture the distinctions between pre-defined different entity categories. The latter part is the attentive conditional random field for decoding to acquire the relationships of recognition results between the parallel tasks. The input texts pass through the former module and obtain the encoding features matched with different entity categories according to the pre-defined entity categories classes. Afterwards the attentive conditional random fields make the model learns the residual value according to the label results of other classes, and uses the attention mechanism to correct the output of the original conditional random fields. Figure~\ref{fig2} illustrates an overview of the model structure.

\subsection{Adaptive Shared Pre-trained Model}
The data representation method follows the paper of Devlin et al.~\cite{devlin2018bert}. We use special ‘[CLS]’ and ‘[SEP]’ tokens at the beginning and end of sentences respectively, and add ‘[PAD]’ tokens at the end of sentences to make their lengths equal to the maximum sequence length. All models are character-level based.

The adaptive shared mechanism~\cite{sun2020adashare} will assign a learnable weight $\alpha_{ij}$ to each transformer encoder layer of the BERT pre-trained model, and update the value of the weight during backward propagation. $i$ indicates that the number of entity categories classes, and the number of transformer encoder layers of the pre-trained model defines the max value of $j$. They can be calculated as:

\begin{equation}
    h_{j+1} = \operatorname{TransformerEncoder}(h_j),
\end{equation}

\begin{equation}
    E_i = \sum^n_{j=0} \alpha_{ij} h_j,
\end{equation}

\noindent where $n$ denotes the total number of transformer encoder layers of the pre-trained language model, $E_i$ denotes the encoder results of $i^{th}$ entity categories classes. In addition, in order to ensure that the weight can better reflect the actual action of each layer after each batch of learning, and prevent the weight from disappearing or exploding, the $softmax$ function is used to output the corrected value of the weight after calculation, as follows:
\begin{equation}
    \alpha_{ij}^* = \operatorname{softmax}(\alpha_{ij}) = \frac{\exp^{\alpha_{ij}}}{\sum^n_{j=0} \exp^{\alpha_{ij}}}.
\end{equation}

\subsection{Attentive Conditional Random Fields}
Conditional random fields (CRF) were proposed by Lafferty et al.~\cite{lafferty2001conditional} in 2001. It has mixed the characteristics of maximum entropy model and hidden Markov model. CRF has been widely used in state-of-the-art NER models to help make better decisions, which considers strong label dependencies by adding transition scores between neighboring labels. \textit{Viterbi} algorithm is applied to search for the label sequence with the highest probability during the decoding process. For $y = \{y_1, ..., y_N \}$ being a sequence of predictions with length $N$. Its score is defined as follows.

\begin{equation}
    s(x, y)=\sum_{i=0}^{N-1} T_{y_{i}, y_{i+1}}+\sum_{i=1}^{N} H_{i, y_{i}},
\end{equation}

\noindent where $T_{y_{i}, y_{i+1}}$ represents the transmission score from $y_i$ to $y_{i+1}$, $H_{i, y_{i}}$ is the score of the $j^{th}$ tag of the $i^{th}$ word from the adaptive shared encoder. CRF model defines a family of conditional probability $p(y|x)$ over all possible tag sequences $y$:

\begin{equation}
    p(y|x)=\frac{\exp ^{s(x, y)}}{\sum_{\tilde{y} \in y} \exp ^{s(x, \tilde{y})}},
\end{equation}

\noindent during training phase, we consider the maximum log probability of the correct predictions. While decoding, we search the tag sequences with maximum score:

\begin{equation}
    y^{*}=\arg \max_{\tilde{y} \in y} \operatorname{score}(x, \tilde{y}).
\end{equation}

Supposed that there are $m$ entity categories classes that we defined in advance. It means that we have $m$ parallel CRFs. For each CRF, the inference results of other parallel CRFs are used as the query of the attention mechanism. Here, set $C$ denotes the \textit{Viterbi} decoding results of all CRFs except the current one, and $d_l$ indicates the max sequence length of each input text. We calculate the attention value of $i^{th}$ CRF as:

\begin{equation}
    Q_i = \sum_{y^*_i \in C} W_f [y^*_{i[0, N]}; c_{(N, d_l]}] + b_f,
\end{equation}

\noindent where $W_f \in \mathbb{R}^{d_{t} \times d_{l}}$, $b_f \in \mathbb{R}^{d_{t} \times d_{l}}$. $c$ denotes the padding constant, and is set to 0 in our paper. $d_t$ denotes the number of tags plus the padding zero. Then we exploit attention mechanism to explicitly learn the dependencies between the origin score and the parallel results and capture the inner structure information of sentence.

\begin{equation}
    R_i = \operatorname{Attention}({Q_i^T}, {K_i}, {V_i})=\operatorname{softmax}\left(\frac{{Q_i^T} {K_i}}{\sqrt{{d_t}}}\right) {V_i}
\end{equation}

\noindent where $K_i \in \mathbb{R}^{d_{t} \times d_{l}}$, $V_i \in \mathbb{R}^{d_{t} \times d_{l}}$ are keys matrix and value matrix, respectively. In our setting, $K_i=V_i=H_i$, denotes that the input of $i^{th}$ CRF. Then we plus the residual $R_i$ and the origin $H_i$ to $i^{th}$ CRF again, and get the final prediction label list.

\begin{equation}
    \begin{split}
            y^{*} &= \arg \max_{\tilde{y} \in y} \operatorname{s}(x, \tilde{y}) \\
            &= \arg \max_{\tilde{y} \in y} \sum_{i=0}^{N-1} T_{y_{i}, y_{i+1}}+\sum_{i=1}^{N} (R+H)_{i, y_{i}}.
    \end{split}
\end{equation}

\section{Experiments}

\subsection{Dataset}

We conduct experiments on the Chinese medical NER, namely the Chinese Medical Entity Extraction dataset (CMeEE)~\cite{hongying2020building}, which is a subtask of Chinese Biomedical Language Understanding Evaluation\footnote{https://tianchi.aliyun.com/cblue} (CBLUE). While absorbing the previous CHIP/CCKS/CCL and other academic evaluation tasks, the dataset also appropriately increases the industry data. Hence, this dataset can well cover the task of Chinese NER in the medical field.

CMeEE dataset contains 20,000 texts and we randomly split these data into training, development, and test sets by 14:3:3, respectively. The goal of this task is to detect and extract the named entities from Chinese medical texts, and divide them into one of the nine pre-defined categories. The statistical results of the nine types of named entities are shown in Table~\ref{tab1}. Dataset provider points that all nested named entities are allowed in the \textit{sym} entity category, and other eight types of entities are allowed inside the entity. So we divide these nine entity categories into two classes: one contains the \textit{sym} entity category and the other one contains eight other entity categories.

\begin{table}[tbhp]
\tabcolsep=0.12cm
\renewcommand\arraystretch{1.4}
\centering
\caption{Category and number of entities in the experimental dataset.}\label{tab1}
\begin{tabular}{c|c|c|c|c|c|c|c|c|c|c}
\hline
{\bfseries Categories} & dis & sym & pro & equ & dru & ite & bod & dep & mic & {\bfseries Total}\\
\hline
{\bfseries Number} & 20,778 & 16,399 & 8,389 & 1,126 & 5,370 & 3,504 & 23,580 & 458 & 2,492 & {\bfseries 82,096}\\
\hline
\end{tabular}
\end{table}

\subsection{Experimental Setup}
Precision ($P$), Recall ($R$) and F-score($F1$) are used to evaluate the predicted entities. An entity is confirmed correct when the predicted category and predicted span both are completely correct. Our model is implemented with PyTorch and we run experiments on NVIDIA Tesla V100 with 32 GB memory.

\begin{table}[tbhp]
\centering
\caption{Architecture hyper-parameters.}\label{tab2}
\begin{tabular}{c|c}
\hline
{\bfseries Hyper-parameters} & {\bfseries Values}\\
\hline
Batch size & 16\\
Embedding size & 768\\
Max sequence length & 128\\
Dropout rate & 0.1\\
LSTM hidden size & 500\\
BERT learning rate & 4e-5\\
ACRF learning rate & 2e-4\\
AdamW weight decay & 1e-5\\
\hline
\end{tabular}
\end{table}

We use the AdamW optimizer for parameter optimization~\cite{loshchilov2017decoupled}. Most of the model hyper-parameters are listed in Table~\ref{tab2}. The pre-trained model used in this paper comes from~\cite{cui2021pre}, namely \textit{BERT-wwm-ext, Chinese}, with 12-layer, 768-hidden, 12-heads and 110M parameters.

\subsection{Results and Comparisons}

Table~\ref{tab3} shows the overall comparisons of different models for the end-to-end Chinese medical NER. We conduct several neural models as baselines. First is the BiLSTM-CRF, which adopts BiLSTM~\cite{huang2015bidirectional} for encoding input texts and a single CRF for decoding entity labels. And BERT-MLP~\cite{cui2021pre}, which employs the linear layer for label classification above the BERT pre-trained language model. Another one is BERT-CRF~\cite{li2020chinese}, which uses CRF for decoding on the basis of the BERT encoding features.

\begin{table*}
\centering
\tabcolsep=0.14cm
\caption{Overall comparisons of different models on CMeEE dataset, while bold ones indicate the best $F_1$ of each categories.}
\label{tab3}
\begin{tabular}{ccccccccccccc}
\toprule
\multirow{2}{*}{Categories} & \multicolumn{3}{c}{BiLSTM-CRF} & \multicolumn{3}{c}{BERT-MLP} & \multicolumn{3}{c}{BERT-CRF} & \multicolumn{3}{c}{ASAC} \\
\cmidrule(r){2-4} \cmidrule(r){5-7} \cmidrule(r){8-10} \cmidrule(r){11-13}
&  $P$  &  $R$  &  $F_1$
&  $P$  &  $R$  &  $F_1$
&  $P$  &  $R$  &  $F_1$
&  $P$  &  $R$  &  $F_1$ \\
\midrule
dis & 56.8 & 55.6 & 56.2 & 63.7 & 64.3 & 64.0 & 63.9 & 64.1 & 64.0 & 64.0 & 65.9 & \textbf{64.9}\\
sym & 38.2 & 37.0 & 37.6 & 47.5 & 51.4 & 49.4 & 49.6 & 51.5 & 50.5 & 52.0 & 50.5 & \textbf{51.5}\\
pro & 40.9 & 40.1 & 40.5 & 53.7 & 51.8 & 52.7 & 54.2 & 56.2 & 55.2 & 55.3 & 56.6 & \textbf{55.9}\\
equ & 50.5 & 24.3 & 32.9 & 65.4 & 64.0 & \textbf{64.7} & 61.2 & 67.7 & 64.3 & 60.7 & 68.8 & 64.5\\
dru & 58.8 & 55.2 & 56.9 & 63.8 & 71.4 & 67.4 & 67.3 & 72.3 & \textbf{69.7} & 66.7 & 71.2 & 68.9\\
ite & 32.6 & 14.1 & 19.7 & 32.1 & 32.0 & 32.0 & 38.6 & 29.0 & 33.1 & 38.8 & 37.6 & \textbf{38.2}\\
bod & 54.4 & 42.5 & 47.7 & 58.9 & 51.8 & 55.1 & 61.4 & 52.6 & 56.6 & 59.7 & 62.9 & \textbf{61.2}\\
dep & 60.0 & 30.9 & 40.8 & 58.1 & 52.9 & \textbf{55.4} & 57.4 & 45.9 & 51.0 & 58.3 & 51.5 & 54.7\\
mic & 68.4 & 56.4 & 61.8 & 67.9 & 68.1 & 68.0 & 66.8 & 67.4 & 67.1 & 67.3 & 69.2 & \textbf{68.2}\\
\hline
Overall & 50.2 & 44.2 & 47.0 & 56.8 & 56.0 & 56.4 & 58.6 & 56.4 & 57.5 & 58.8 & 60.3 & \textbf{59.5}\\
\bottomrule
\end{tabular}
\end{table*}

On the CMeEE dataset, our method improves the $F_1$ scores by 12.5, 3.1, and 2.0 respectively, compared with BiLSTM-CRF, BERT-MLP, and BERT-CRF. Furthermore, our approach achieves better results in 6 of the 9 entity categories, surpassing other baseline models.

\subsection{Ablation Studies}
In this paper, we introduce the interactions of adaptive shared mechanism and attentive conditional random fields to respectively help better predict nested named entities in the dataset. We implement an ablation study to verify the effectiveness of the interactions. We conduct four experiments: (a) our model with adaptive shared mechanism and attentive conditional random fields. (b) without adaptive shared mechanism: we skip Eq. (2)-(3) and only used the final output layer of the BERT-based pre-trained model for encoding. (c) without attentive conditional random fields: we skip Eq. (7)-(9) and use separate CRFs to predict the labels of different category classes, and combine the results. (d) without adaptive shared mechanism and attentive conditional random fields: a combination of (b) and (c), different from BERT-CRF, there are two independent CRFs for decoding entities from two category classes. Table~\ref{tab4} shows the experimental results. Experimental results show that adaptive shared mechanism and attentive conditional random fields both make a positive effect on prediction, while attentive conditional random fields are a little more effective. Compared to abandoning the above two methods, our model received a 1.12 F1 score boost.

\begin{table*}[tbhp]
\tabcolsep=0.14cm
\centering
\caption{Ablation study on CMeEE dataset.}
\label{tab4}
\begin{tabular}{l|ccc}
\toprule
&  $P$  &  $R$  &  $F_1$\\
\midrule
Ours            &  \textbf{58.78} & \textbf{60.30} & \textbf{59.53}\\
Ours(w/o AS)    &  58.38 & 59.74 & 59.05\\
Ours(w/o ACRF)  &  58.14 & 59.73 & 58.89\\
Ours(w/o AS,ACRF)& 58.59 & 58.24 & 58.41\\
\bottomrule
\end{tabular}
\end{table*}
\vspace{-0.2in}

\subsubsection{Analysis of the Adaptive Shared Mechanism.}
\begin{figure}
\includegraphics[width=\textwidth]{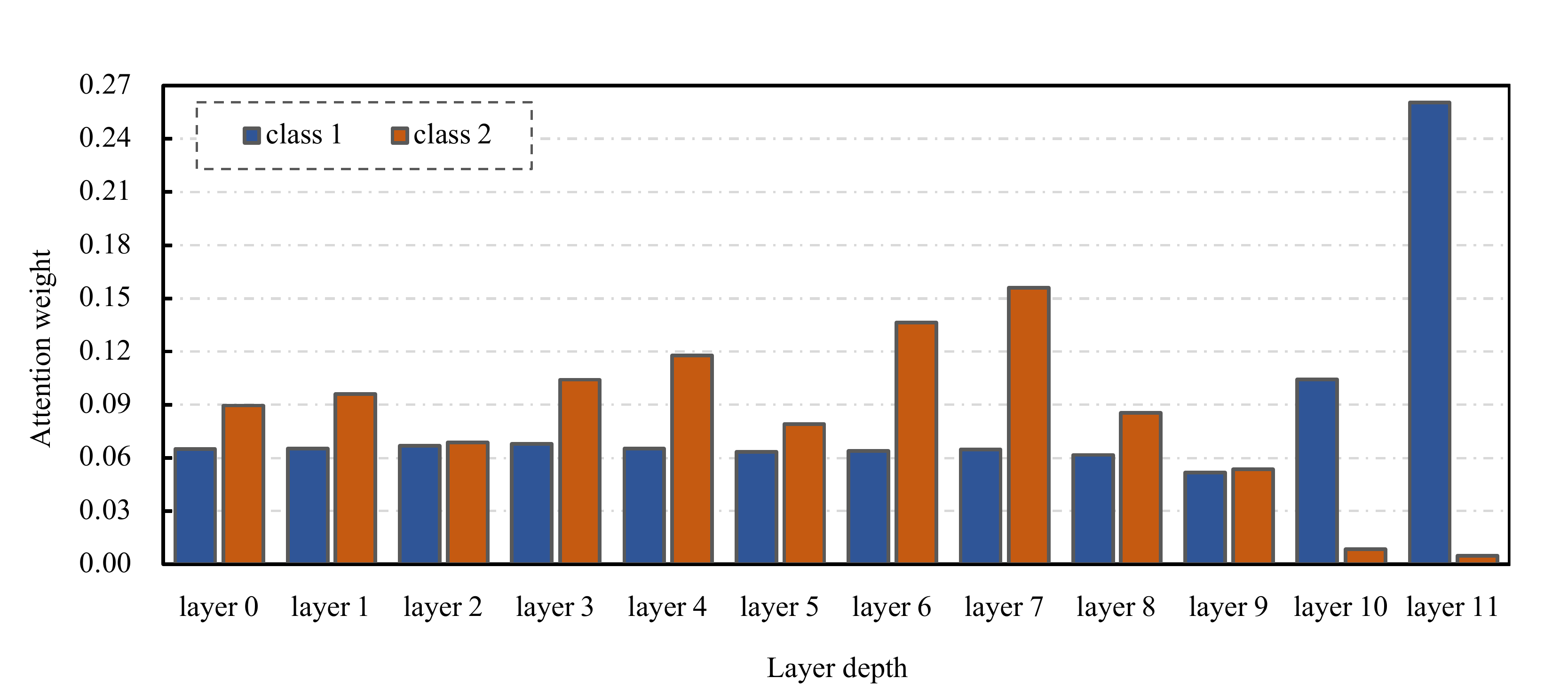}
\caption{The adaptive shared weights of different layers and entity category classes of the pre-trained model.} \label{fig3}
\end{figure}

Some researches~\cite{jawahar2019does} show that different context features are stored in different layers of the BERT-based pre-trained language model. Therefore, we reasonably infer that different features of hidden layers can variously affect the recognition of different entity categories. As mentioned above, we divide entity categories into two classes. Under the adaptive shared mechanism, there are two groups of weights: $\{\alpha_{1,0}, ... ,\alpha_{1,11}\}$ for class 1, $\{\alpha_{2,0}, ... ,\alpha_{2,11}\}$ for class 2 to extract the different layers of the pre-trained model. Figure~\ref{fig3} shows the weights of two classes after training.

According to the data shown in the figure, we notice that class 1 prefers the upper layers outputs of the BERT-based model while middle layers outputs have a greater impact on class 2. Therefore, the adaptive shared mechanism enables the model to learn the context features that match the nested entity category better, which is conducive to subsequent decoding.

\subsubsection{Analysis of the Attentive Conditional Random Fields.}
In the decoding module, we count the changed labels between the outputs of the initial CRF and the outputs plus the residual of the CRF after the attention mechanism. Table~\ref{tab5} evaluates the performance of attentive conditional random fields. In a total of 155,658 tokens of 3,000 test cases, 4,816 and 1,392 prediction labels for the two classes have changed, where the positive changes are exceeded 60\%.

\begin{table}[!t]
\renewcommand\arraystretch{1.1}
\tabcolsep=0.14cm
\center
\caption{Statistics of labels positive changed after attentive conditional random fields.}\label{tab5}
\begin{tabular}{c|cc}
\hline
& Class 1 & Class 2\\
\hline
Changed & 4,816 & 1,392 \\
Positive & 3,170 & 857 \\
Positive ratio & 65.8\% & 61.5\% \\
\hline
\end{tabular}
\end{table}

\section{Conclusion}
In this paper, we proposed a novel end-to-end model ASAC for Chinese nested named entity recognition in the medical domain, which can adaptively learn the distinctions and connections between different entities recognition tasks based on sequence labeling. First, we introduce the adaptive shared mechanism to get features of different layers of the BERT-based model to different nested entity category classes. Moreover, by constructing attentive conditional random fields, our model can exploit the encoding features to decode corresponding labels by blending other predictions via the attention mechanism. Extensive experiments conducted on publicly available datasets show the effectiveness of the proposed method. In the future, we will apply this novel model to more different types of datasets to verify validity.

\subsubsection{Acknowledgements.}
This research was partially supported by grants from the National Natural Science Foundation of China (Grant No. 61922073) and the Joint Fund for Medical Artificial Intelligence (Grant No. MAI2022C007).
%
%
%
\bibliographystyle{splncs04}
\bibliography{ref}
%




\end{document}